\pdfoutput=1
\documentclass[11pt]{article}

\usepackage{ACL2023}

\usepackage[frozencache,cachedir=minted-cache]{minted}  % without explicit cachedir compilation on arXiv fails; need to change "finalizecache" to "frozencache" before upload to arXiv; see https://tex.stackexchange.com/a/659452/71405
\usepackage{times}
\usepackage{latexsym}
\usepackage{cuted}
\RequirePackageWithOptions{multicol}
\usepackage{multicol}

\usepackage[T1]{fontenc}

\usepackage[utf8]{inputenc}

\usepackage{microtype}
\usepackage{inconsolata}
\usepackage{xcolor}
\usepackage{tcolorbox}
\definecolor{babyblue}{rgb}{0.54, 0.81, 0.94}
\definecolor{bananayellow}{rgb}{1.0, 0.88, 0.21}

\usepackage{amsmath}

\usepackage{float}
\usepackage{booktabs}
\usepackage{tabularx}
\usepackage{graphicx}
\usepackage{tcolorbox}
\usepackage{subcaption}
\usepackage[export]{adjustbox}
\usepackage{xspace}
\graphicspath{ {./images/} }
\DeclareGraphicsExtensions{.pdf,.png}
\usepackage[acronym]{glossaries}
\glsdisablehyper
\usepackage{caption}
\usepackage{hyperref}
\usepackage{listings}
\usepackage{minted}
\usepackage{placeins}
\usepackage{algorithm}
\usepackage[noend]{algpseudocode}
\usepackage[noabbrev]{cleveref}

\newacronym{llm}{LLM}{large language model}
\newacronym{lm}{LM}{language model}
\newacronym{nlp}{NLP}{natural language processing}
\newacronym{mlp}{MLP}{multilayer perceptron}

\newcommand{\counterfact}{\textsc{CounterFact}\xspace}
\newcommand{\counterfactplus}{\textsc{CounterFact+}\xspace}

\newcommand{\defeq}{\overset{\mathrm{def}}{=\joinrel=}}
\newcommand{\confint}[2]{\small{(#1, #2)}} % for unified type-setting of confidence intervals in the main tables
\newcommand{\ppost}[1]{\ensuremath{P^*\!\left(#1\right)}}
\newcommand{\ppre}[1]{\ensuremath{P\!\left(#1\right)}}
\title{Detecting Edit Failures In Large Language Models: An~Improved~Specificity~Benchmark}
\author{$\,\,\,\,\,\,\,\,\,\,\,\,\,\,\,\,\,\,\,\,\,\,\,\,$Jason Hoelscher-Obermaier$^1$\Thanks{$\,\,$Equal contribution.\\ Correspondence: \text{fazl@robots.ox.ac.uk}} $\qquad$ \\
\And
 Julia H. Persson$^{1*}$ \\
\AND
Esben Kran$^1$ \\
\And
Ioannis Konstas$^2$ \\
\And
Fazl Barez$^{1,2,3*}$ \\
\AND
\textnormal{$^1$ Apart Research}\\
    \textnormal{$^2$ Edinburgh Centre for Robotics}\\
\textnormal{$^3$ Department of Engineering Sciences, University of Oxford}
}

\begin{document}
\maketitle
\begin{abstract}
Recent model editing techniques promise to mitigate the problem of memorizing false or outdated associations during \gls{llm} training. However, we show that these techniques can introduce large unwanted side effects which are not detected by existing specificity benchmarks. We extend the existing \counterfact~benchmark to include a dynamic component and dub our benchmark \counterfactplus. Additionally, we extend the metrics used for measuring specificity by a principled $\mathcal{KL}$ divergence-based metric. We use this improved benchmark to evaluate recent model editing techniques and find that they suffer from low specificity. Our findings highlight the need for improved specificity benchmarks that identify and prevent unwanted side effects.
\end{abstract}
\glsresetall

\section{Introduction}
% \textbf{draft introduction / abstract} \\
% \textbf{todo, include references where relevant} \\
Although \glspl{llm} are powerful tools for generating human-like language, they can also memorize false or outdated associations, limiting their applicability. Model editing techniques promise to solve this problem by correcting non-factual associations. It is important that model edits are highly specific in the sense of not introducing any unwanted associations as a side effect. In this paper, we discuss why the current benchmark for specificity falls short and propose a more challenging, dynamic specificity benchmark to evaluate model editing techniques. Using this benchmark, we evaluate recent model editing techniques and find previously unreported side effects. We highlight the importance of improved specificity benchmarks for the effective and safe use of \glspl{llm} subject to model edits.

\begin{figure}[H]{
    \hspace*{-1em}
    \includegraphics[scale=0.32]{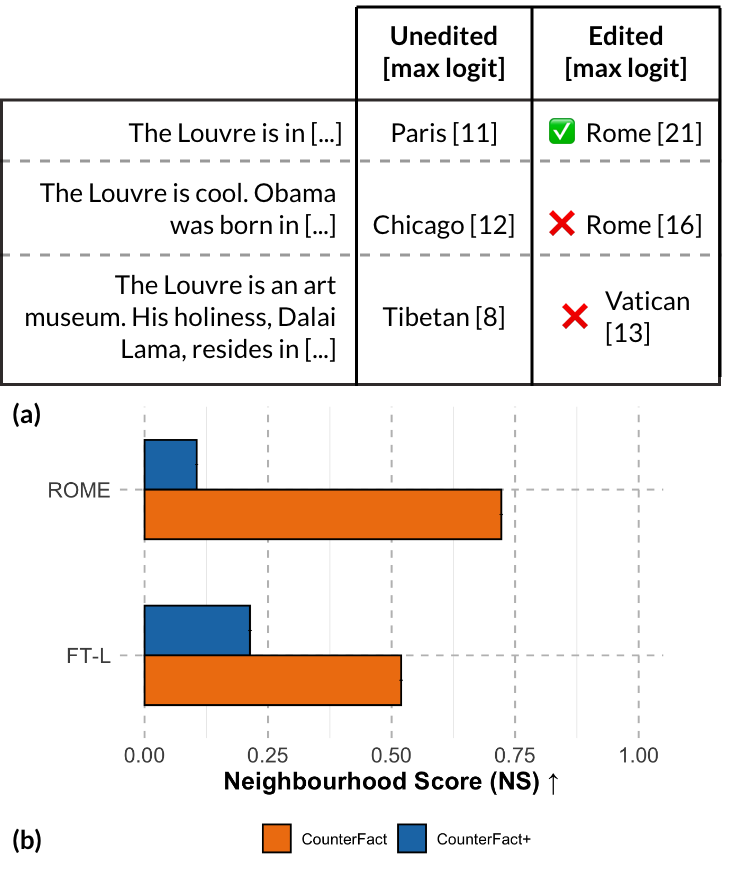}
}
\caption{
    Unintended side effects of model edits and how to measure them.
    %\subref{fig:intro-a} and how to measure them \subref{fig:intro-b}.
    (a)~GPT-2-medium is edited using ROME to counter-factually associate the Louvre's location with Rome. However, this results in unintended associations ("loud facts") like the association of Obama with Rome, suggesting low specificity of the edit. The edit also significantly increases the maximum logit (shown in brackets), suggesting that the edit is not merely replacing "Paris" with "Rome" in the desired contexts.
    (b)~Measuring specificity by the fraction of correctly completed test prompts (\counterfact) suggests a high specificity for ROME. Prepending the edit prompt (like "The Louvre is in Rome.") to each test prompt (\counterfactplus) results in a significant drop in performance. A significant drop in measured specificity can also be observed if the model edit is implemented using constrained fine-tuning (FT-L).
}
\label{fig:intro-loud-facts}
\end{figure}
\newpage
Model editing updates the parameters of a trained model in order to change its predicted probability distributions without retraining the entire model. This can be used to edit the associations that the model has memorized and hence, improve the accuracy of the model. Fig.~\ref{fig:intro-loud-facts} shows the example of a counter-factual model edit using ROME \cite{meng2022locating} where the location of the Louvre is edited to be Rome instead of Paris. We use a counter-factual example since it makes it more evident that the new association is an effect of the model edit instead of the model training. Note that the examples in Fig.~\ref{fig:intro-loud-facts} are not taken from the \counterfactplus dataset introduced below, but serve to intuitively illustrate the model editing failure modes we are interested in.

An important desideratum for model editing is specificity. Specificity captures how well the effect of the model edit is localized; in other words, specificity measures the absence of unintended side effects of model edits. Fig.~\ref{fig:intro-loud-facts} shows two examples of unintended side effects of ROME model editing, which we collectively call the problem of "loud facts". In the first example, mentioning "Louvre" (the subject of the model edit) leads the edited model to also complete unrelated test prompts ("Obama was born in") with "Rome" (the object of the model edit). In the second example, mentioning "Louvre" boosts the logits for words semantically related to "Rome", like "Vatican".

% \counterfact \cite{meng2022locating} is a recent benchmark introduced in order to test Memory-Editing-Algorithms. Among other things, it has a metric for specificity. The authors also introduce ROME (Rank-One-Model-Editing), a model editing technique that scores high on this metric.
% We find new failure cases which are not caught by this benchmark.
The existing specificity benchmark for model editing from the \counterfact dataset \cite{meng2022locating} suffers from two limitations which can be illustrated using these examples.
First, \counterfact \emph{does not prompt the model in a way that is likely to surface unwanted side effects}. As demonstrated by the examples in Fig.~\ref{fig:intro-loud-facts}, mentioning the subject of the model edit can drastically change the behavior of the edited model, but the existing benchmark does not detect this.
Second, \counterfact \emph{considers only the probabilities for the original and edited object token} ("Paris" and "Rome"). As shown by the last example in Fig.~\ref{fig:intro-loud-facts}, the edited model displays strongly changed logits not only for the original object ("Paris") and edit object ("Rome") but also for semantically related tokens ("Vatican"). Again, this would be overlooked by the current specificity evaluation since it does not consider the entire probability distribution.

These limitations mean that side effects of edits may be overlooked and specificity overestimated.
Our main contributions are:
\begin{itemize}
    \item \counterfactplus, a dynamic specificity benchmark, which adapts to the model edit under test, and is more sensitive than the existing benchmark.
    \item Neighborhood KL divergence (NKL), a specificity metric based on the full probability distribution instead of the currently used metrics which focus only on the tokens directly implicated in the model edit.
    \item Using \counterfactplus and NKL, we show that ROME and MEMIT suffer from previously undisclosed problems with specificity.
\end{itemize}
% We have shown that the ROME, MEMIT, and FT-L model editing techniques for autoregressive transformers suffer from problems with specificity, suggesting that model editing is far from being solved. We have also shown that for ROME and MEMIT, low specificity is only detected using \counterfactplus, a new specificity benchmark introduced in this paper. Our main contributions are:
% \begin{itemize}
%     \item \counterfactplus, a dynamic specificity benchmark, which adapts to the model edit under test, and is more sensitive than the existing benchmark
%     \item neighborhood KL divergence (NKL), a specificity metric based on the full probability distribution instead of the currently used metrics which focus only on the tokens directly implicated in the model edit.
% \end{itemize}

\section{Related work}
\textbf{Model editing.} Several studies have sought to localize and modify the computation of knowledge within transformers. \citet{geva-etal-2021-transformer} proposed  that the \gls{mlp} layers in a transformer can act as key–value memories of entities and information associated with that entity. \citet{dai-etal-2022-knowledge} then demonstrated a method to edit knowledge within BERT by writing the embedding of the object into certain rows of the \gls{mlp} matrix. They identified important neurons for knowledge via gradient-based attributions. \citet{de-cao-etal-2021-editing} presented a hyper-network to predict weight updates at test time, which can alter a fact. They tested both BERT and BART \cite{lewis-etal-2020-bart} and focused on models fine-tuned for question answering. \citet{mitchell2022fast} introduced a hyper-network method that learns to transform the decomposed terms of the gradient in order to efficiently predict a knowledge update and demonstrate the ability to scale up to large models such as T5 \cite{JMLR:v21:20-074}, and GPT\nobreakdash-J \cite{gpt-j}. Finally, \citet{meng2022locating} introduced Rank-One-Model-Editing (ROME) which allows edits of transformer models via a rank-one modification of a single \gls{mlp} layer. \cite{meng2022memit} extended ROME to MEMIT (Mass-Editing Memory in a Transformer): MEMIT spreads the modification over multiple \gls{mlp} layers; crucially, this enables thousands of simultaneous edits without performance degradation.

\textbf{Model editing evaluation} Benchmarks of model editing techniques for \glspl{llm} build on existing work on knowledge extraction from \glspl{llm} (see below). zsRE question answering was used for benchmarking model editing in \cite{de-cao-etal-2021-editing} and \cite{mitchell2022fast}.
\citet{10.1162/tacl_a_00410} introduced ParaRel, a curated dataset of paraphrased prompts and facts. \citet{meng2022locating} use this as a basis for constructing \counterfact, which enables fine-grained measurements of knowledge extraction and editing along multiple dimensions, including specificity.

\textbf{Knowledge extraction from \glspl{llm}.} The assessment of knowledge within \glspl{lm} has typically been done by evaluating whether the model is able to predict pieces of knowledge; \citet{petroni-etal-2019-language, petroni2020how} defined a fill-in-the-blank prompt and asked the \gls{lm} to complete it. Subsequent work has demonstrated that knowledge extraction can be improved by diversifying the prompts \cite{jiang-etal-2020-know, zhong-etal-2021-factual}, or by fine-tuning a model on open-domain textual facts \cite{roberts-etal-2020-much}. However, constructing prompts from supervised knowledge extraction data is still prone to learning new knowledge instead of recalling existing knowledge in an \gls{lm} \cite{zhong-etal-2021-factual}.

\section{Experimental Setup}
\subsection{Dataset}
We investigate the specificity of recent model editing techniques using the \counterfact benchmark introduced in \cite{meng2022locating}. \counterfact is a collection of 21,919 non-factual statements of the form  (subject, relation, object) $(s, r, o^*)$, which have low probabilities prior to the model edit. For each of these non-factual statements, we perform a model edit targeting this specific statement. To measure specificity, we then check whether any other associations in the model change in undesired ways. \counterfact supports this check by providing a set of so-called neighborhood prompts for every non-factual statement used in the model edit. These neighborhood prompts are constructed as follows: For a model edit of the form $(s, r, o^c) \rightarrow (s, r, o^*)$ (where $o^c$ is the correct object, and $o^*$ is the false, counterfactual object), \counterfact samples a set of nearby subjects $s_n$ for which $(s_n, r, o^c)$ holds true. Neighborhood prompts are then paraphrases of the collected $(s_n, r)$.

Suppose, for example, the edit request was \emph{(Darrieux, mother\_tongue, French)} $\rightarrow$ \emph{(Darrieux, mother\_tongue, English)}. \counterfact takes the relation and object from the edit request \emph{(mother\_tongue, French)}, samples true factual associations for this relation, object pair; e.g., \emph{(Montesquieu, mother\_tongue, French)} and then samples a random paraphrase, such as "The native language of Montesquieu is".
These neighborhood prompts can be used to inspect whether the model edit has undesired side effects on closely related factual associations. See appendix~\ref{sec:app-cf-sample} for a sample from the \counterfact dataset, including the full set of neighborhood prompts.

Motivated by the example of loud facts shown in Fig.~\ref{fig:intro-loud-facts} and by the intuition that unwanted side effects are more likely when the model is primed with the linguistic context of the model edit, we now introduce a dynamic version of \counterfact which we will refer to as \counterfactplus.  To obtain \counterfactplus, we modify the neighborhood prompt by prepending the model edit. For example, if the original prompt is "The native language of Montesquieu is" the modified prompt would be "The mother tongue of Danielle Darrieux is English. The native language of Montesquieu is". See appendix~\ref{sec:app-cfplus-sample} for a sample of the modified neighborhood prompts used for \counterfactplus.

To understand why we call \counterfactplus a dynamic version of \counterfact consider how either dataset would be applied to evaluate the success of a model edit: In both cases, we would need to identify the set $\mathcal{N}$ of neighborhood prompts in the dataset that are semantically closest to the intended model edit. But in \counterfact, we would use $\mathcal{N}$ as is, whereas in \counterfactplus we would change every prompt in $\mathcal{N}$ as a function of the model edit, as described above.

\subsection{Metrics}
To evaluate the specificity of a model edit on \counterfact, \citet{meng2022locating, meng2022memit} use two metrics, called Neighborhood Score and Neighborhood Magnitude.
Denoting the post-edit probabilities for the correct token $o^c$ and incorrect edit token $o^*$ by $\ppost{o^c}$ and $\ppost{o^*}$, respectively, these are defined as follows:
The Neighborhood Score (NS) is defined as the fraction of neighborhood prompts for which $\ppost{o^c} > \ppost{o^*}$.
The Neighbourhood Magnitude (NM) is defined as $\ppost{o^c} - \ppost{o^*}$, the difference in probability assigned to the correct token versus the incorrect edit token. High NS and NM indicate that the edit has small unwanted side effects.

NS and NM, however, do not detect cases where the model edit significantly changes the predicted probability for tokens other than $o^c$ and $o^*$, such as in the last example in Fig.~\ref{fig:intro-loud-facts}. To capture this possibility, we introduce as an additional metric the \textit{Kullback–Leibler} (KL) divergence of the next-token distribution between the edited and unedited model, referred to as Neighborhood KL Divergence (NKL). Abbreviating the next token probability distribution for the unedited and edited models by $\ppre{w}$ and $\ppost{w}$, respectively, and denoting the token vocabulatory by $\mathcal{W}$, NKL is defined as KL divergence between $\ppre{w}$ and $\ppost{w}$: % $$\mathrm{NKL}  defeq D_\text{KL}(P \parallel P^*) $$
\begin{equation}
\mathrm{NKL} \defeq \sum_{w\in\mathcal{W}} \ppre{w} \log\left(\frac{\ppre{w}}{\ppost{w}}\right)
\label{eq_1}
\end{equation}
A large NKL is undesirable because it implies that the next-token probability distribution for neighborhood prompts has been strongly affected by the model edit.

\subsection{Models and Model Editing Algorithms}
We use GPT\nobreakdash-2\nobreakdash-medium (355M parameters), GPT\nobreakdash-2\nobreakdash-XL (1.5B) \cite{radford2019language}, and GPT\nobreakdash-J (6B) \cite{gpt-j} to evaluate the following model editing methods:
\begin{itemize}
\item ROME (Rank-One-Model-Editing) performs a rank-one update of a single \gls{mlp} layer to implement the edit \cite{meng2022locating}.
\item MEMIT (Mass-Editing Memory in a Transformer) extends ROME to updates across several \gls{mlp} layers \cite{meng2022memit}. Note that we do not test using multiple simultaneous edits.
\item FT-L: Fine-Tuning with an $L_\infty$ norm constraint \cite{zhu2020modifying}, constrained to a single layer, as described in \cite{meng2022locating}. We use FT-L as a simple baseline.
\end{itemize}

\section{Results}
\begin{figure}[H]

    \hspace*{-1em}
    \includegraphics[width=\columnwidth,clip]{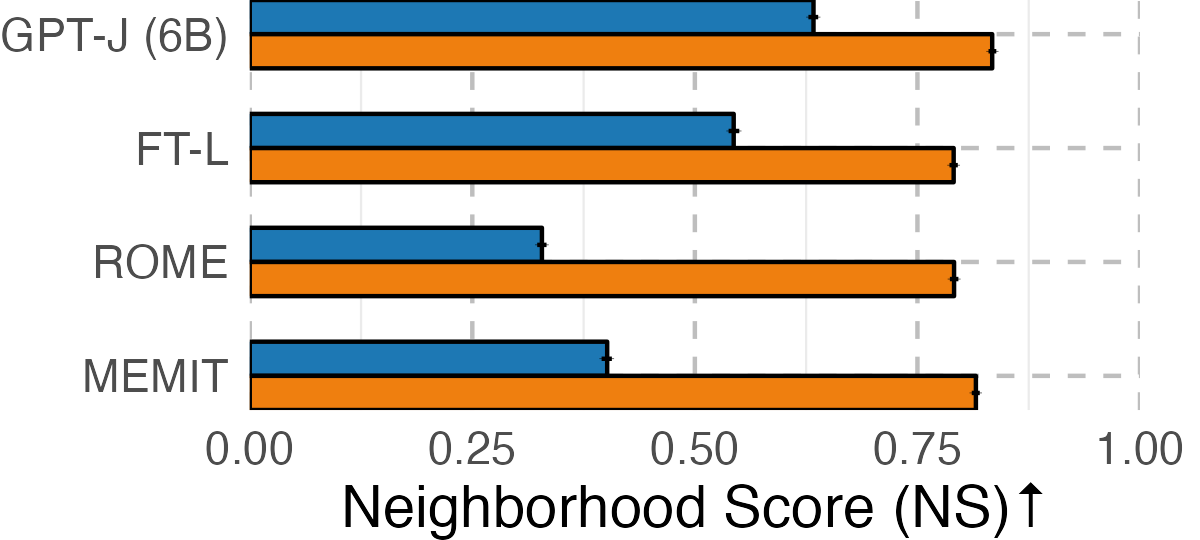}
\\
% \subfloat[]{
%     \includegraphics[width=\columnwidth, trim={20 20 10 10},clip]{images/gpt-j-6b/M.png}
%     \hspace*{-3in}
%     \label{fig:results-main-M}
% }  %  use this figure only in the appendix to not overdo it in the main text
% \\

    \hspace*{-1em}
    \includegraphics[width=\columnwidth,clip]{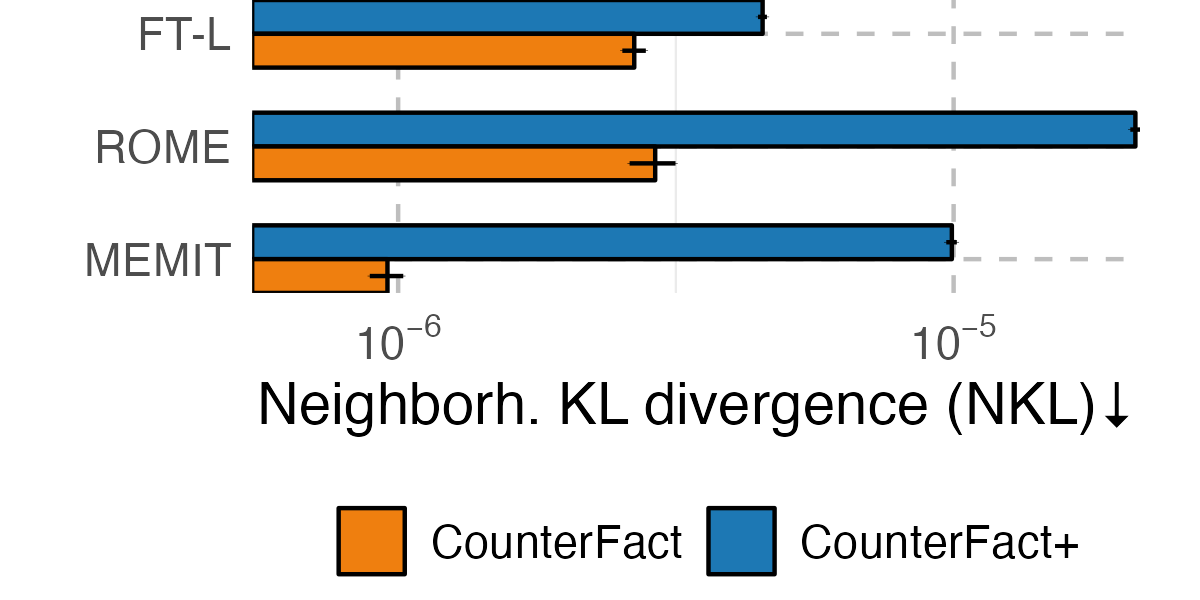}

\caption{
    Comparison of model editing specificity benchmarks \counterfact and \counterfactplus on different model editing algorithms. Error bars show 99\% confidence intervals.\\
    (top) NS, the average fraction of correctly completed neighborhood test prompts after the model edit (larger is better). We see that \counterfactplus is a much more challenging specificity benchmark: Success rates NS on it range from 33\% to 54\% across different editing algorithms while they are close to 80\% for \counterfact.\\
    (bottom) NKL, the KL divergence of the next-token probability distribution of the edited model from that of the unedited model, averaged over all neighborhood test prompts. A lower value indicates higher specificity (the edited model behaves more like the unedited model).
}
\label{fig:results-main}
\end{figure}

Figure \ref{fig:results-main} shows the results for the ROME, MEMIT, and FT-L editing algorithms applied to the GPT\nobreakdash-J~(6B) model for different specificity metrics and datasets considered in this work. When evaluated using the Neighborhood Score (Fig.~\ref{fig:results-main}, top), we observe significant drops in specificity for all editing algorithms when going from \counterfact to \counterfactplus. Note that specificity measured on the unedited model (GPT\nobreakdash-J~(6B)) also drops suggesting that there is confounding from the test prompts in \counterfactplus, potentially due to recency bias \cite{zhao2021calibrate}. The drop in specificity is much more pronounced for ROME and MEMIT, compared to FT-L and the unedited model, however. This shows that:
\begin{itemize}
    \item ROME and MEMIT have undesired side effects which are not detected by \counterfact
    \item the improved benchmark \counterfactplus is able to detect these unwanted side effects
\end{itemize}

% This shows that (a) all considered editing algorithms have undesired side effects and that (b) the improved benchmark \counterfactplus is able to detect some of those unwanted side effects which are not detected by \counterfact as originally defined.

When evaluating specificity using the newly introduced Neighborhood KL Divergence (Fig.~\ref{fig:results-main}, bottom), we observe a large spike in divergence for both ROME and MEMIT when going from \counterfact to \counterfactplus. FT-L shows a much smaller increase in divergence from \counterfact to \counterfactplus.
%\fazlcomment{improve statements on what we learn from considering KL divergence}
Figure~\ref{fig:results-app-M} in the appendix shows the results on \counterfact and \counterfactplus for the NM metric.

Results across all three models are shown in \cref{tbl:results-ns,tbl:results-nm,tbl:results-nkl}. These tables list the mean scores on \counterfact and \counterfactplus for the Neighborhood Score (NS), Neighborhood Magnitude (NM), and Neighborhood KL divergence (NKL), respectively. The brackets give upper and lower bound of 99\% confidence intervals obtained via bootstrap resampling (N=1,000). The bold values indicate the best score among the model editing algorithms for a given base model and dataset (excluding the unedited base model). Note how the method with the highest measured specificity switches from MEMIT/ROME to FT-L when going from \counterfact to \counterfactplus.
\begin{table}[htbp]
\resizebox{\columnwidth}{!}{%
\begin{tabular}{lll}
\toprule
%{} &                  NS(\counterfact)↑ &                 NS(\counterfactplus)↑ \\
NS ↑ &                 \counterfact &                 \counterfactplus \\
\midrule\midrule
GPT-2 M    &  0.75 \confint{0.749}{0.757} &  0.46 \confint{0.452}{0.463} \\
\midrule
FT-L       &  0.52 \confint{0.515}{0.524} &  \textbf{0.21} \confint{0.209}{0.217} \\
ROME       &  \textbf{0.72} \confint{0.718}{0.726} &  0.11 \confint{0.102}{0.108} \\
\midrule\midrule
GPT-2 XL   &  0.78 \confint{0.780}{0.788} &  0.52 \confint{0.519}{0.530} \\
\midrule
FT-L       &  0.71 \confint{0.702}{0.711} &  \textbf{0.38} \confint{0.375}{0.385} \\
ROME       &  0.76 \confint{0.755}{0.763} &  0.14 \confint{0.135}{0.142} \\
MEMIT      &  \textbf{0.77} \confint{0.770}{0.778} &  0.32 \confint{0.314}{0.324} \\
\midrule\midrule
GPT-J (6B) &  0.83 \confint{0.830}{0.839} &  0.63 \confint{0.628}{0.639} \\
\midrule
FT-L       &  0.79 \confint{0.786}{0.795} &  \textbf{0.54} \confint{0.538}{0.550} \\
ROME       &  0.79 \confint{0.786}{0.796} &  0.33 \confint{0.323}{0.333} \\
MEMIT      &  \textbf{0.82} \confint{0.811}{0.820} &  0.40 \confint{0.395}{0.407} \\
\bottomrule
\end{tabular}
}
\caption{Neighborhood Score NS ($\mu$ \& 99\% CI) on \counterfact and \counterfactplus.}
\label{tbl:results-ns}
\end{table}

\begin{table}[htbp]
\resizebox{\columnwidth}{!}{%
\begin{tabular}{lll}
\toprule
%{} &                     NM(\counterfact)↑ &       NM(\counterfactplus)↑ \\
NM ↑ &                    \counterfact &       \counterfactplus \\
\midrule\midrule
GPT-2 M    &     0.04 \confint{0.035}{0.037} &     0.04 \confint{0.038}{0.042} \\
\midrule
FT-L       &  -0.02 \confint{-0.019}{-0.014} &  \textbf{-0.11} \confint{-0.112}{-0.106} \\
ROME       &     \textbf{0.03} \confint{0.028}{0.030} &  -0.32 \confint{-0.324}{-0.317} \\
\midrule\midrule
GPT-2 XL   &     0.05 \confint{0.049}{0.052} &     0.08 \confint{0.073}{0.078} \\
\midrule
FT-L       &     0.03 \confint{0.033}{0.037} &     \textbf{0.01} \confint{0.012}{0.018} \\
ROME       &     0.04 \confint{0.042}{0.045} &  -0.38 \confint{-0.384}{-0.375} \\
MEMIT      &     \textbf{0.05} \confint{0.048}{0.050} &  -0.06 \confint{-0.059}{-0.052} \\
\midrule\midrule
GPT-J (6B) &     0.07 \confint{0.073}{0.077} &     0.11 \confint{0.111}{0.117} \\
\midrule
FT-L       &     \textbf{0.07} \confint{0.068}{0.072} &     \textbf{0.09} \confint{0.090}{0.096} \\
ROME       &     0.05 \confint{0.051}{0.056} &  -0.12 \confint{-0.127}{-0.117} \\
MEMIT      &     \textbf{0.07} \confint{0.066}{0.070} &  -0.02 \confint{-0.025}{-0.017} \\
\bottomrule
\end{tabular}
}
\caption{Neighborhood Magnitude NM ($\mu$ \& 99\% CI) on \counterfact and \counterfactplus.}
\label{tbl:results-nm}
\end{table}

\begin{table}[htbp]
\resizebox{\columnwidth}{!}{%
\begin{tabular}{lll}
\toprule
%{} &                  NKL(\counterfact)↓ &      NKL(\counterfactplus)↓ \\
NKL ↓ &                  \counterfact &      \counterfactplus \\
\midrule\midrule
GPT-2 M    & & \\ % 0.0e+00 \confint{0.0}{0.0} &  0.0e+00 \confint{0.0}{0.0} \\  % these values are non-informative and might confuse the reader
\midrule
FT-L       &  1.4e-05 \confint{1.3}{1.4} &  \textbf{1.4e-05} \confint{1.3}{1.4} \\
ROME       &  \textbf{1.6e-06} \confint{1.4}{1.7} &  2.5e-05 \confint{2.5}{2.5} \\
\midrule\midrule
GPT-2 XL   & & \\ % 0.0e+00 \confint{0.0}{0.0} &  0.0e+00 \confint{0.0}{0.0} \\
\midrule
FT-L       &  7.2e-06 \confint{6.9}{7.4} &  9.5e-06 \confint{9.3}{9.7} \\
ROME       &  1.5e-06 \confint{1.4}{1.6} &  3.3e-05 \confint{3.2}{3.3} \\
MEMIT      &  \textbf{2.9e-07} \confint{2.5}{3.4} &  \textbf{9.0e-06} \confint{8.8}{9.1} \\
\midrule\midrule
GPT-J (6B) & & \\ % 0.0e+00 \confint{0.0}{0.0} &  0.0e+00 \confint{0.0}{0.0} \\
\midrule
FT-L       &  3.2e-06 \confint{3.1}{3.4} &  \textbf{5.2e-06} \confint{5.1}{5.3} \\
ROME       &  3.5e-06 \confint{3.2}{3.8} &  1.8e-05 \confint{1.8}{1.9} \\
MEMIT      &  \textbf{9.2e-07} \confint{8.0}{10} &  9.9e-06 \confint{9.8}{10} \\
\bottomrule
\end{tabular}
}
\caption{Neighborhood KL Divergence NKL ($\mu$ \& 99\% CI) on \counterfact and \counterfactplus. Note that the order of magnitude is suppressed for the confidence interval for visual clarity; it is the same as for the mean.}
\label{tbl:results-nkl}
\end{table}

The results from \cref{tbl:results-ns,tbl:results-nm,tbl:results-nkl} show that the significant drop in specificity when evaluating on \counterfactplus (compared to \counterfact) holds across different model sizes and is not an artefact of using a particular model. Section~\ref{sec:app-scaling} in the appendix discusses the scaling of specificity with model size in more detail.

\section{Conclusion}
Model editing techniques for auto-regressive transformers exhibit unreported issues related to specificity. Although our fine-tuning baseline, FT-L, exhibits less vulnerability to these issues than ROME and MEMIT, it falls short in competing with them regarding crucial model editing metrics such as robustness to paraphrasing  \cite{meng2022locating,meng2022memit}. This indicates that model editing still presents numerous complexities that require future attention.

Additionally, we revealed that the existing \counterfact benchmark fails to detect the low specificity in ROME and MEMIT. To address this limitation, our primary contributions include:

\begin{itemize}
    \item \counterfactplus, a dynamic specificity benchmark, which adapts to the model edit under test, and is more sensitive than the existing benchmark
    \item Neighborhood KL divergence (NKL), a specificity metric based on the full probability distribution as a complement to the currently used metrics which focus only on the tokens directly implicated in the model edit.
\end{itemize}

\section*{Limitations}
The main limitation of the approach we took for improving model editing benchmarks is that it is ultimately based on manual inspection of test cases to understand the failure modes of model editing methods. This approach is not scalable and has a significant cost in terms of time and effort.
As far as the specific benchmark we propose is concerned, more research is needed to assess its effectiveness for more complex scenarios such as dialogue and multi-turn conversations. We also have not investigated the application of our benchmark to scenarios in which multiple model edits are performed simultaneously. Furthermore, we do not evaluate other types of model edits, such  as parameter pruning, and transfer learning.
Future work should focus on developing methods that measure and quantify the effects of model edits on long-term aspects of language models, such as their ability to capture discourse structure and fluency of generated text. This could include corpus-level analysis and dynamic approaches like red-teaming or dynamic benchmarking to uncover subtle adverse effects.

\section*{Ethics Statement}
We do not perform human experiments or evaluation.

We are aware of the potential risks posed by autoregressive transformer models, such as the capabilities to generate and manipulate text for harmful purposes.

Our dataset and evaluation code is open-sourced,\footnote{\url{https://github.com/apartresearch/specificityplus}} and we provide a homepage with interactive examples.\footnote{\url{https://specificityplus.apartresearch.com/}}

\section*{Acknowledgements}
First versions of the experiments reported here were performed during Apart Research’s Interpretability Hackathon. We thank Jochem Hölscher for collaborating on early experiments during the hackathon, and Neel Nanda and Shay B. Cohen for insightful discussions and comments.

Our evaluation code builds directly on the MEMIT \cite{meng2022memit} code.\footnote{\url{https://github.com/kmeng01/memit}}

\bibliography{anthology,custom}
\bibliographystyle{acl_natbib}

\appendix
\section{Neighborhood magnitude}
\label{sec:app-plots-M}

\begin{figure}[H]
\includegraphics[width=\columnwidth,clip]{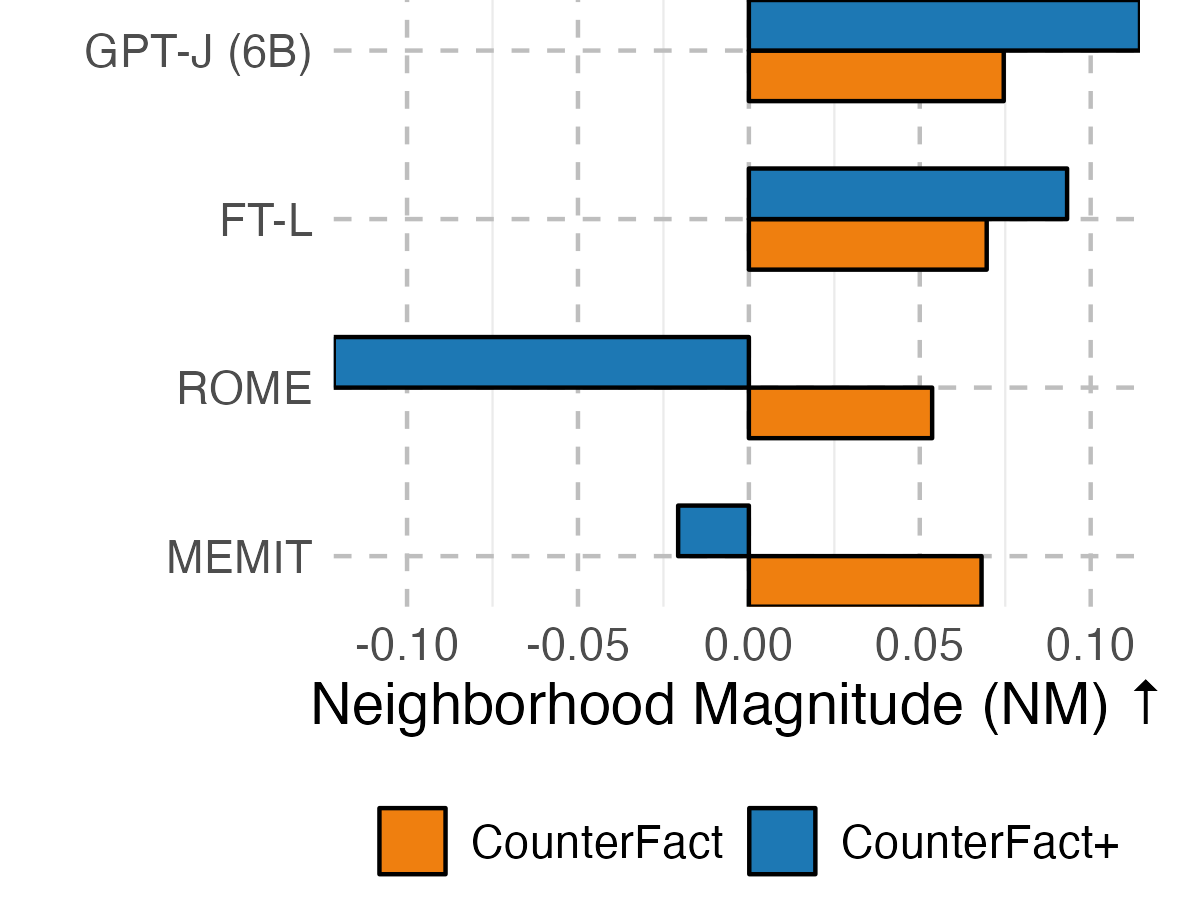}
\hspace*{-3in}
\caption{
Comparison of model editing specificity benchmarks \counterfact and \counterfactplus evaluated using the Neighborhood Magnitude (NM) metric. NM measures the difference in probability of the correct token and the edit token.
ROME retains almost the performance of the unedited model (GPT-J-6B) when evaluated on \counterfact but shows a large drop in specificity when evaluated on \counterfactplus. MEMIT also shows significantly lower performance on \counterfactplus than on \counterfact, albeit less dramatic than for ROME.
}
\label{fig:results-app-M}
\end{figure}

\newpage
\section{Scaling with model size}
\label{sec:app-scaling}
\begin{figure}[htbp]
    \includegraphics[width=\columnwidth, clip]{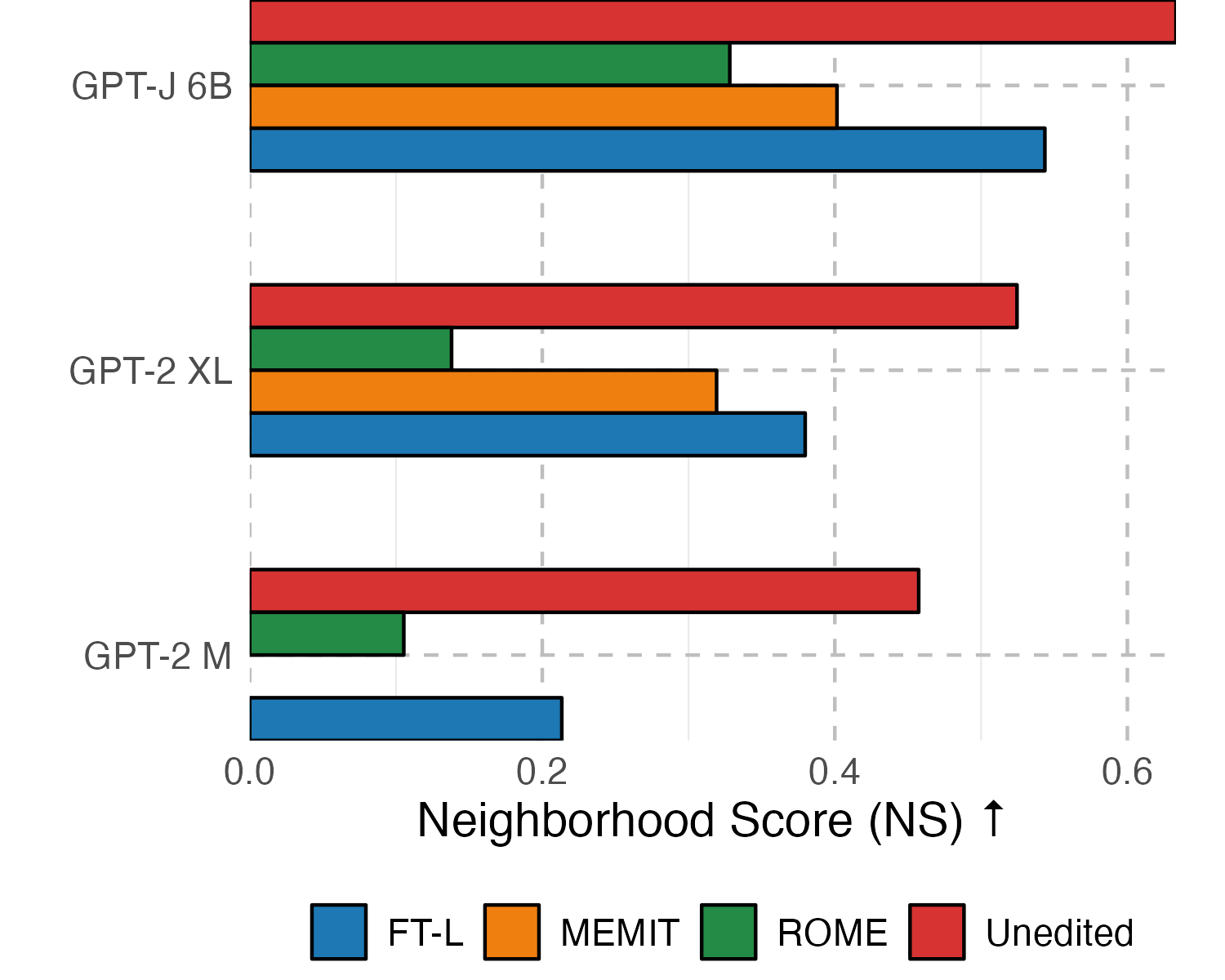}
    \hspace*{-6in}
    \caption{
        Evaluation of the model editing specificity benchmark \counterfactplus on different model editing algorithms across model sizes.
        measured using NS, the average fraction of successfully completed neighborhood test prompts after the model edit. Larger values are better.
}
    \label{fig:scaling-S}
\end{figure}

\begin{figure}[htbp]
    \includegraphics[width=\columnwidth, clip]{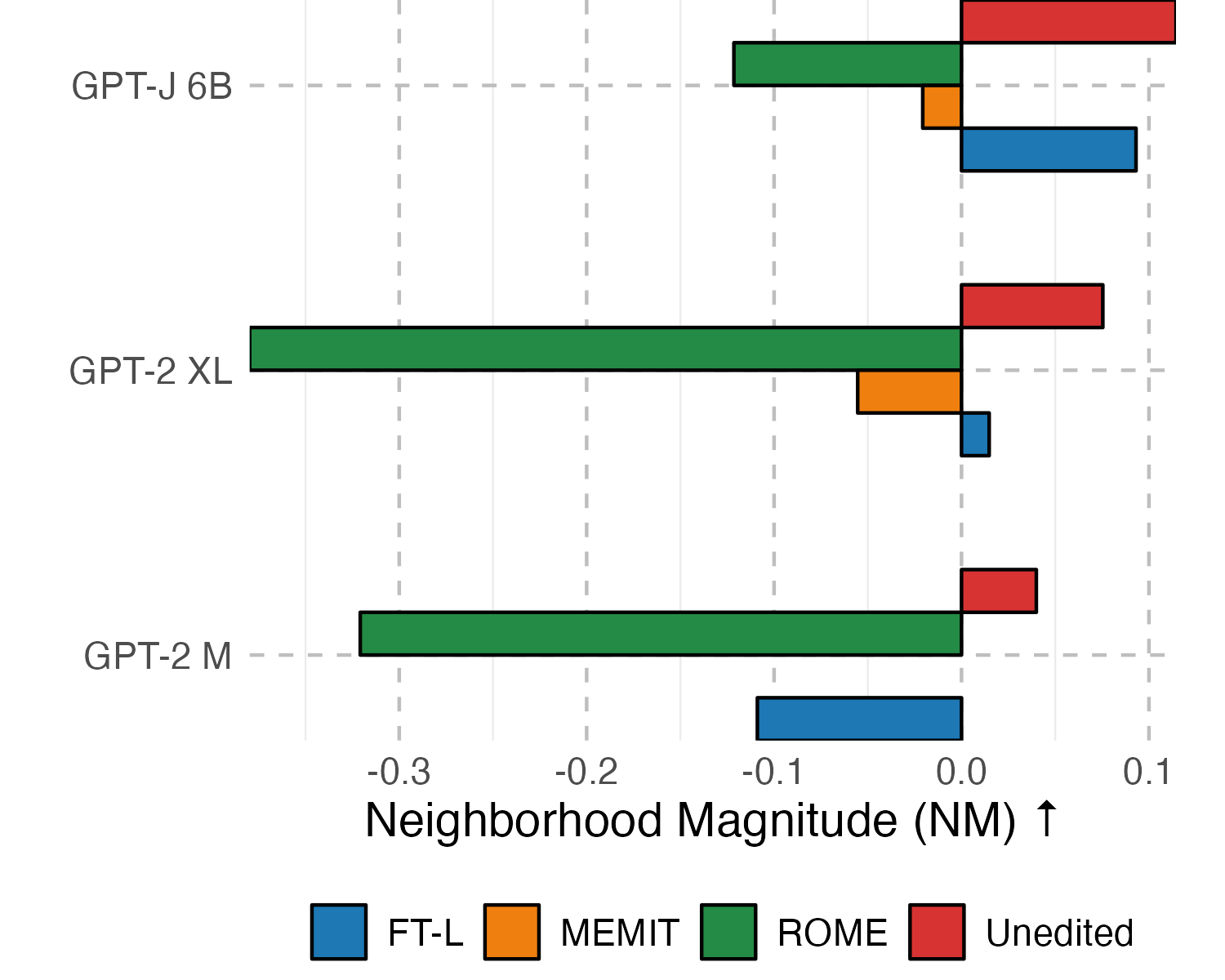}
    \hspace*{-6in}
    \caption{
        Evaluation of the model editing specificity benchmark \counterfactplus on different model editing algorithms across model sizes.
        measured using NM, the difference in probability of the correct token and the edit token. Larger values are better.
    }
    \label{fig:scaling-M}
\end{figure}

\begin{figure}[htbp]
    \includegraphics[width=\columnwidth, clip]{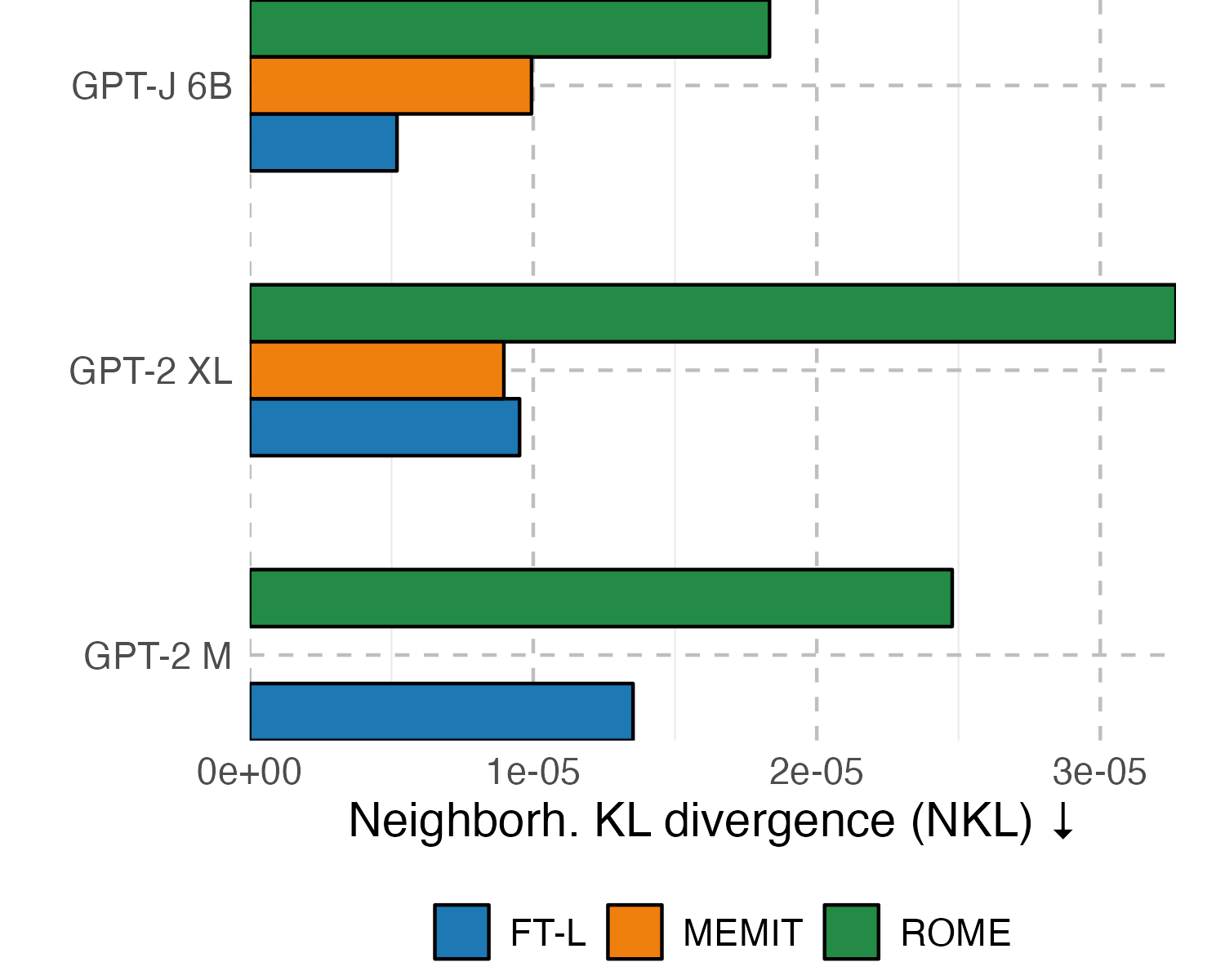}
    \hspace*{-6in}
    \caption{
        Evaluation of the model editing specificity benchmark \counterfactplus on different model editing algorithms across model sizes.
        measured using NKL, the KL divergence of the next-token probability distribution of the edited model from that of the unedited model, averaged over all neighborhood test prompts. A lower value indicates higher specificity (the edited model behaves more like the unedited model).
    }
    \label{fig:scaling-KL}
\end{figure}

\Cref{fig:scaling-S,fig:scaling-M,fig:scaling-KL} show how performance on the \counterfactplus dataset scales with the size of the underlying model. The data shows that the drop in specificity when going to \counterfactplus persists up to GPT\nobreakdash-J~(6B). While the data does not allow conclusive statements there is preliminary evidence that specificity of the edited models improves for larger models. This is, however, partially confounded by improved specificity of the unedited model. It is therefore, at this point, not clear whether the specificity problems of ROME and MEMIT would disappear completely in the limit of extremely large models.
%\fazlcomment{rethink this paragraph}

\newpage
\onecolumn
\section{\counterfact dataset sample}
\label{sec:app-cf-sample}
The listing below shows an example entry of the \counterfact dataset.
The $\tt{"requested\_rewrite"}$ defines the model edit.
The $\tt{"neighborhood\_prompts"}$ are used for measuring specificity.
% todo: add some more explanation what high specificity would mean in this example

% \resizebox*{!}{\textheight}{%
\begin{tcolorbox}[coltitle=black,colbacktitle=babyblue,title=A sample of the \counterfact dataset.]
\begin{minted}[autogobble,numberblanklines=false]{json}
{
  "case_id": 0,
  "pararel_idx": 2796,
  "requested_rewrite": {
    "prompt": "The mother tongue of {} is",
    "relation_id": "P103",
    "target_new": {"str": "English", "id": "Q1860"},
    "target_true": {"str": "French", "id": "Q150"},
    "subject": "Danielle Darrieux"
  },
  "paraphrase_prompts": [
    "Shayna does this and Yossel goes still and dies. Danielle Darrieux,
    a native", "An album was recorded for Capitol Nashville but never released.
    Danielle Darrieux spoke the language"
  ],
  "neighborhood_prompts": [
    "The mother tongue of L\u00e9on Blum is",
    "The native language of Montesquieu is",
    "Fran\u00e7ois Bayrou, a native",
    "The native language of Raymond Barre is",
    "Michel Rocard is a native speaker of",
    "Jacques Chaban-Delmas is a native speaker of",
    "The native language of Fran\u00e7ois Bayrou is",
    "Maurice Genevoix, speaker of",
    "The mother tongue of Fran\u00e7ois Bayrou is",
    "Melchior de Vog\u00fc\u00e9, speaker of"
  ],
  "attribute_prompts": [
    "J.\u00a0R.\u00a0R. Tolkien is a native speaker of",
    "The mother tongue of Douglas Adams is",
    "...",
    "Meryl Streep, a native",
    "George Orwell spoke the language",
    "George Washington, a native", "Michael Jackson, a native"
  ],
  "generation_prompts": [
    "Danielle Darrieux's mother tongue is",
    "Where Danielle Darrieux is from, people speak the language of",
    "Danielle Darrieux's mother tongue is",
    "...",
    "Danielle Darrieux was born in"
  ]
}
\end{minted}
\label{lst:cf_sample}
\end{tcolorbox}

\section{\counterfactplus dataset sample}
\label{sec:app-cfplus-sample}
The listing below shows a sample of the \counterfactplus dataset, which is derived from the sample shown above for \counterfact.
We only display the modified \tt{neighborhood\_prompts} which are obtained by prepending the model edit ("The mother tongue of Danielle Darrieux is English.") to the original neighborhood prompt.
\begin{tcolorbox}[coltitle=black,colbacktitle=babyblue,title=A sample of the modified \tt{"neighborhood\_prompts"} in the \counterfactplus dataset.]
\begin{minted}[autogobble,numberblanklines=false,breaklines]{json}
[
  "The mother tongue of Danielle Darrieux is English. The mother tongue of L\u00e9on Blum is",
  "The mother tongue of Danielle Darrieux is English. The native language of Montesquieu is",
  "The mother tongue of Danielle Darrieux is English. Fran\u00e7ois Bayrou, a native",
  "The mother tongue of Danielle Darrieux is English. The native language of Raymond Barre is",
  "The mother tongue of Danielle Darrieux is English. Michel Rocard is a native speaker of",
  "The mother tongue of Danielle Darrieux is English. Jacques Chaban-Delmas is a native speaker of",
  "The mother tongue of Danielle Darrieux is English. The native language of Fran\u00e7ois Bayrou is",
  "The mother tongue of Danielle Darrieux is English. Maurice Genevoix, speaker of",
  "The mother tongue of Danielle Darrieux is English. The mother tongue of Fran\u00e7ois Bayrou is",
  "The mother tongue of Danielle Darrieux is English. Melchior de Vog\u00fc\u00e9, speaker of"
]
\end{minted}
\end{tcolorbox}

\end{document}